\documentclass[lettersize,journal]{IEEEtran}
\usepackage{amsmath,amsfonts}
\usepackage{algorithmic}
\usepackage{array}
\usepackage[caption=false,font=normalsize,labelfont=sf,textfont=sf]{subfig}
\usepackage{textcomp}
\usepackage{stfloats}
\usepackage{url}
\usepackage{verbatim}
\usepackage{graphicx}
\usepackage{amssymb}
\usepackage{booktabs}
\usepackage{slashbox}
\usepackage{multirow}
\usepackage{makecell}
\usepackage{ragged2e}
\usepackage[pagebackref,breaklinks,colorlinks]{hyperref}
\def\BibTeX{{\rm B\kern-.05em{\sc i\kern-.025em b}\kern-.08em
    T\kern-.1667em\lower.7ex\hbox{E}\kern-.125emX}}
\usepackage{balance}
\begin{document}
\title{Transferring Dual Stochastic Graph Convolutional Network for Facial Micro-expression Recognition}
\author{Hui Tang, Li Chai and Wanli Lu
\thanks{This paper was supportted by the National Natural Science Foundation of China under grants (61800339)}}

\markboth{Journal of \LaTeX\ Class Files, March~2022}%
{How to Use the IEEEtran \LaTeX \ Templates}

\maketitle

\begin{abstract}
Micro-expression recognition has drawn increasing attention due to its wide application in lie detection, criminal detection and psychological consultation. To improve the recognition performance of the small micro-expression data, this paper presents a transferring dual stochastic Graph Convolutional Network (TDSGCN) model. We propose a stochastic graph construction method and dual graph convolutional network to extract more discriminative features from the micro-expression images. We use transfer learning to pre-train SGCNs from macro expression data. Optical flow algorithm is also integrated to extract their temporal features. We fuse both spatial and temporal features to improve the recognition performance. To the best of our knowledge, this is the first attempt to utilize the transferring learning and graph convolutional network in micro-expression recognition task. In addition, to handle the class imbalance problem of dataset, we focus on the design of focal loss function. Through extensive evaluation, our proposed method achieves state-of-the-art performance on SAMM and recently released MMEW benchmarks. Our code will be publicly available accompanying this paper.
\end{abstract}

\begin{IEEEkeywords}
Class, IEEEtran, \LaTeX, paper, style, template, typesetting.
\end{IEEEkeywords}

\section{Introduction}
\label{Intro}

Facial expression is an effective and universal way of expressing human emotions and intentions \cite{Corneanu2016}. There are two kinds of facial expression, daily macro-expression and hidden micro-expression. Micro-expressions are genuine and involuntary emotions which people usually attempt to control and conceal under high-stake situations. As the development of artificial intelligence, micro-expression recognition attracts more and more attentions due to its importance in lie detection, criminal detection and psychological consultation \cite{Ekman2003, Bhushan2015, Ekman2009}.

Micro-expression recognition is a complicated and challenging task in the field of computer vision \cite{Ben2021, Xie2020, He2017MER}. The available databases are usually small. The biggest dataset consists only 300 samples. Their intensity is low and duration is short. Instead of a single image, image sequences are necessary in the micro-expression recognition task.


One of the most efficient ways is to extract more discriminative features from image sequences. Traditional methods utilize handcrafted features to deal with the micro-expression recognition problem, including local binary patterns (LBP) \cite{Ojala2002}, LBP on three orthogonal planes (LBP-TOP) \cite{Zhao2007}, and histogram of oriented gradients (HOG) \cite{Davision2015}. These methods are restricted by the limited training samples. Their performance need to be further improved. With the success of deep learning in many computer vision applications, deep features also obtain state-of-the-art predictive performance in micro-expression recognition. Recently, Ben \emph{et al}. \cite{Ben2021} gave a comprehensive and systematic survey of the major challenges and developments in micro-expression recognition area. They also presented a new dataset with more samples called \emph{micro-and-macro expression warehouse} (MMEW). State-of-the-art method for the micro-expression recognition is the transferring long-term convolutional neural network (TLCNN) \cite{Wang2018}, which extracted both spatial and temporal features by feeding the CNN features to Long Short Term Memory (LSTM). The recognition accuracy on MMEW is 69.4\% and SAMM is 73.5\%. This detection and recognition is improved but still limited by the lack of large datasets.

We observe that most of existing methods extract facial features on the basis of pixels. However, face pixels may not reveal the distinct feature of micro-expression images since its salient feature is low intensity and transit.

\begin{figure}[htp]
   \centering
   \includegraphics[width=1\linewidth]{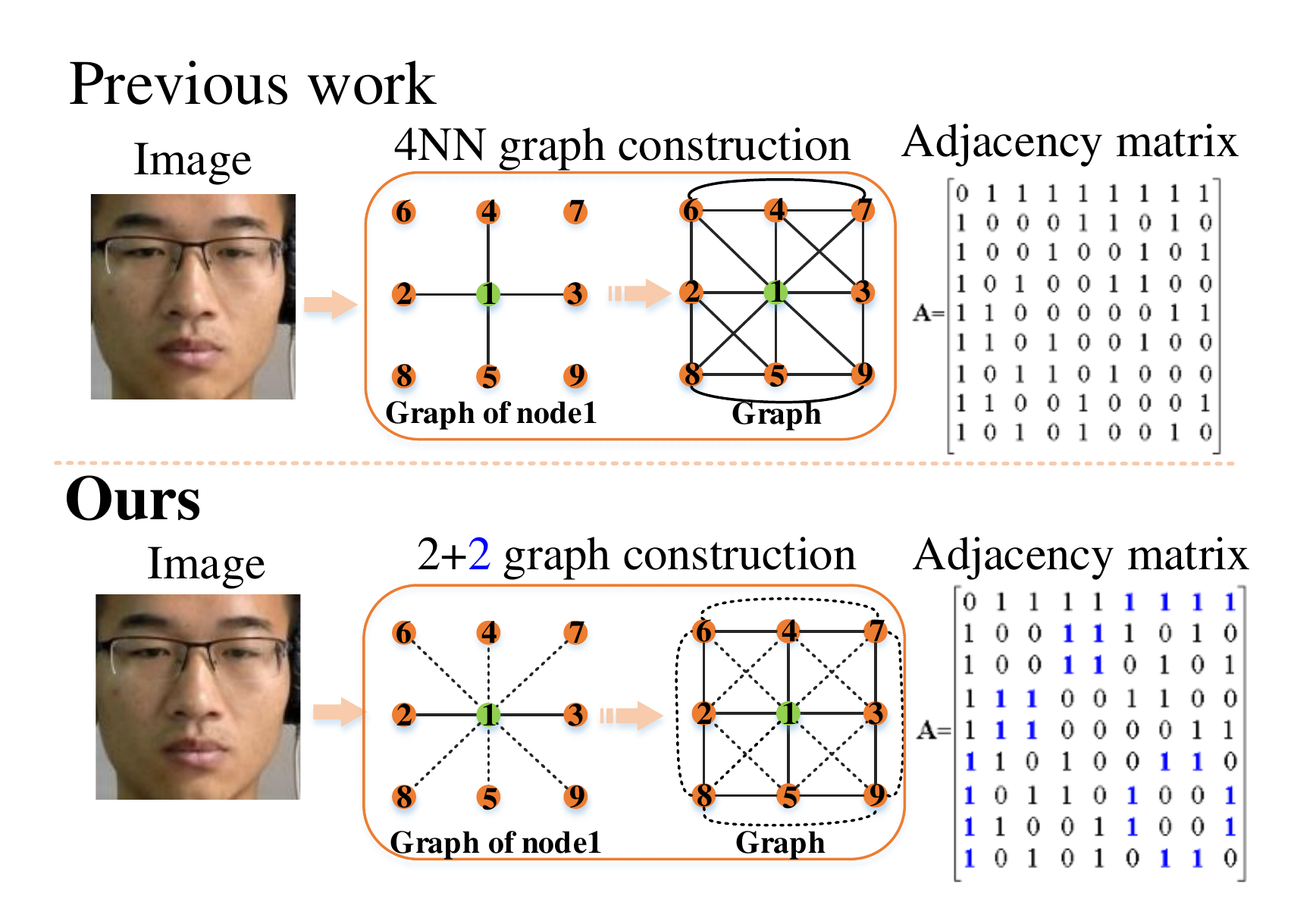}
   \caption{(\textbf{Top}): Previous work often construct graph by their nearest neighbors (NN) to handle the image data. Nodes have edges among the nearest neighbors. Each node has 4 edges in the 4-NN graph. (\textbf{Bottom}): Our work proposes a stochastic $p+q$ graph construction to improve the discriminative learning. Nodes have edges among the $p$ nearest neighbors and $q$ stochastic neighbors. Each node has 2 nearest neighbors and 2 stochastic neighbors in the $2+2$ graph. The potential nodes for stochastic edges are linked in dotted line and represented in blue 1's in the adjacency matrix.}
  \label{stochasticgraph}
\end{figure}

Inspired by the outstanding performance of Graph Convolutional Network (GCN) on both grid and non-grid data \cite{Xu2017, NIPS2016, Landrieu2018, Chen2018}, we propose a transferring stochastic dual GCN (TSDGCN) model to improve the performance of micro-expression recognition. Transfer learning is efficient to tackle the small data problem. We are the first to integrate the GCN and transfer learning to deal with the small data problem in micro-expression recognition. To enhance the discriminative learning, we present a novel graph construction method. We propose a dual graph convolutional network (DGCN), based on which two feature extractors are employed. We observe that previous GCN work on grid data often construct graph by their nearest neighbors. Nodes have edges among the nearest neighbor nodes. This results in redundant links and can be computationally expensive. We propose a stochastic graph construction method to extract geometric feature between pixels named $p+q$ graph construction. Nodes have edges among the $p$ nearest neighbors and $q$ neighbors stochastically selected. There is a simple example of $2+2$ graph construction shown in Figure \ref{stochasticgraph}. With the same number of neighbors, the adjacency matrix of our method may encode broader and more distinctive information, which benefits from the technique of stochastic sampling.

The class imbalance problem is a common problem in the existing micro-expression datasets. Class imbalance seriously decreases the recognition performance. Instead of cross entropy, this paper attempts to design the focal loss function to deal with the problem of class imbalance.


To conclude, the main contributions of this paper are as follows:
\begin{itemize}

\item We propose a novel stochastic GCN model to improve the performance of micro-expression recognition. SGCN can extract both local and global features with less computational complexity.
\item To the best of our knowledge, we are the first to propose a dual GCN architecture to extract more representative features.
\item Our proposed method can outperform the state of the art results on four benchmarks.

\end{itemize}

\begin{figure*}[htp]
   \centering
   \includegraphics[width=0.85\linewidth]{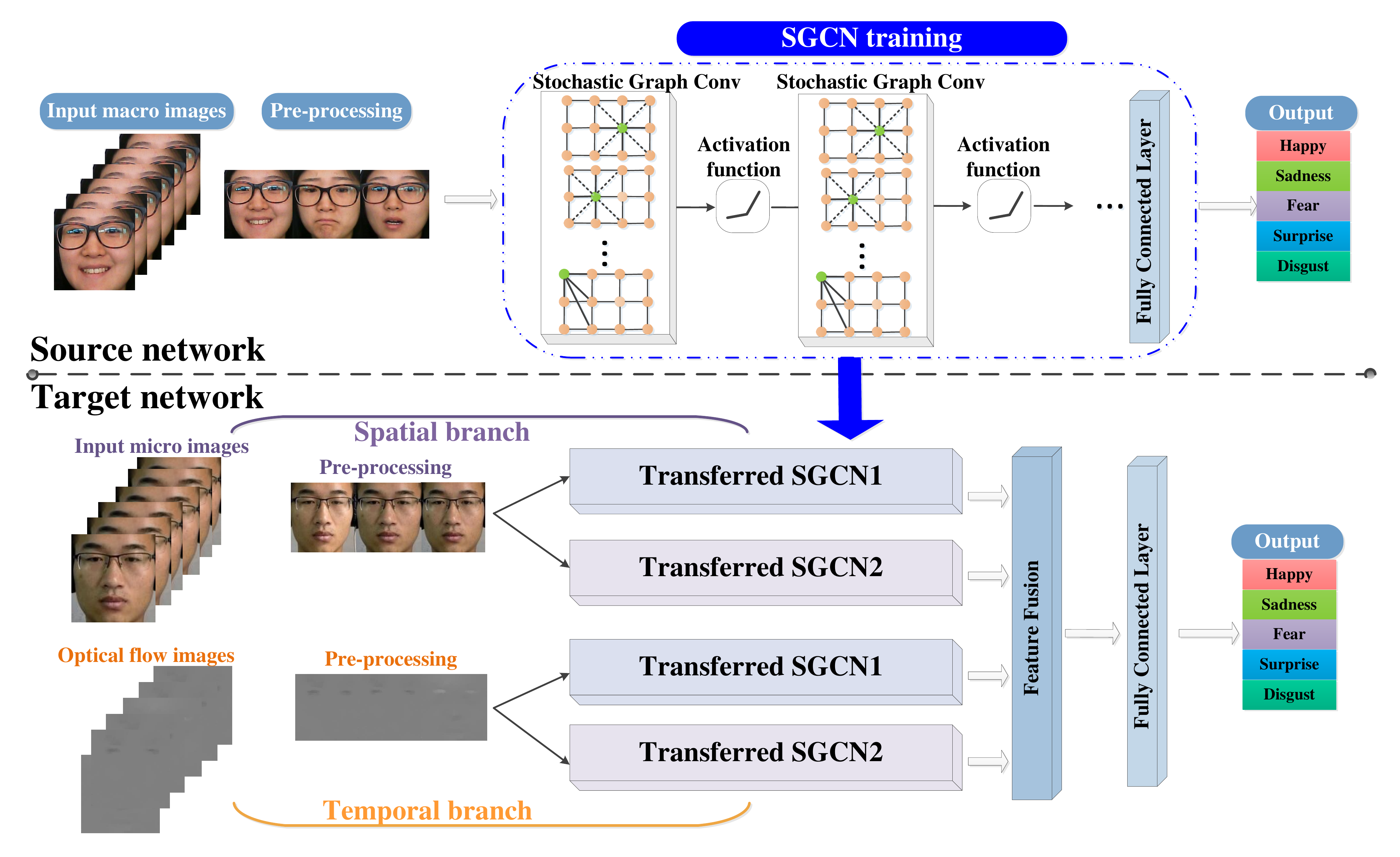}
   \caption{An overview of the proposed network - Transferring Stochastic Dual Graph Convolutional Network}
  \label{figTGCN}
\end{figure*}

\section{Related work}

\textbf{Micro-expression recognition}
Traditional classification methods, like support vector machine (SVM) \cite{Wang2015TIP}, extreme learning machine (ELM) \cite{Wang2014ECCV} and K nearest neighbor \cite{Chaudhry2009} \emph{et al}.,  rely on artificially designed features, which are simple to use but with poor performance. Deep learning methods can automatically extract an optimal feature representation \cite{Patel2017, Peng2017, Khor2018}. Khor \emph{et al}. \cite{Khor2018} presented an Enriched Long-term Recurrent Convolutional Network (ELRCNN) to enrich the subtle movements. However, all existing micro-expression databases are small. Deep learning with small data may not achieve good performance. Transfer learning shows great potential to deal with the small data problem, which uses knowledge from a related domain (in which large datasets are available) \cite{Ojala2002, Zong2018TL, Zong2019}. Most of the existing transfer learning methods are CNN-based architectures. Just like CNN can capture the most significant information within pixels in images, a graph-based learning algorithm learns the relation between each node from the data described in the form of graph. To the best of our knowledge, this is the first work that integrates GCN and transfer learning to handle the micro-expression recognition problem.

\textbf{Graph convolutional network}
Recently, it has been shown that GCN approaches are powerful for extracting discriminative geometric features in both gird and non-grid data application \cite{Pan2021}.  There are two mainstreams to define a GCN, the spectral methods \cite{Bruna2014, NIPS2016, Kipf2017} and the spatial methods \cite{Alessio2009, Atwood2016, Niepert2016, Gilmer2017}. The spectral domain models the representation in the Fourier domain based on eigen-decomposition. The spatial method directly implements operators on the graph node and its neighbors. Up to now, there are few published GCN related work to deal with micro-expression problem. Lo \emph{et al}. \cite{Lo2020} used 3D ConvNets to extract actions units (AUs) features and applied GCN layers to discover the relations between AU nodes for micro-expression recognition.  Xie \emph{et al}. \cite{Xie2020} exploited AUs relational information and proposed AU-assisted Graph Attention Convolutional Network for micro-expression recognition. We propose a novel graph construction method by stochastic sampling and we employ two GCN feature extractors to graph data. 

\textbf{Optical flow}
The idea of optical flow was first introduced by Horn \emph{et al}. \cite{Horn1981} to describe the movement of brightness patterns in an image. The basic concept is to find the distance of an identical object in different frames. By utilizing the pixel-wise difference between consecutive frames in a video can thus be obtained \cite{Zach2007}. Owing to the fact that optical flow could capture temporal patterns between consecutive frames, one of the most employed architecture is to combine optical flow feature with CNN to further recognize spatial patterns \cite{Xia2020, Gan2019}. On one side, we integrate the optical flow feature and SDGCN to extract temporal features. On the other side, optical flow information is also utilized in the data pre-processing to select meaningful frames in the entire micro-expression sequences.

\textbf{Focal loss}
Class imbalance is a common problem in .  Li and Deng \cite{Li2017, Li2019} used local preservation loss to maintain the locality of each examples, making the local neighborhood in each class as compact as possible. Mollahosseini \emph{et al}. \cite{Ali2019} and Ji \emph{et al}. \cite{Ji2020} respectively used the weighted cross entropy loss function to improve the problem of class imbalance by weighting the loss function of each class. The imbalance problem in the field of expression is still difficult to solve. Lin \emph{et al}. \cite{He2017} proposed the definition of focal loss to handle the class imbalance problem of binary classification in the target detection scene of one stage.  It down-weights the
loss assigned to well-classified examples. This paper introduces the focal loss to deal with the problem of class imbalance in micro-expression recognition.

\section{Approach}

\subsection{Overview}

As shown in Figure \ref{figTGCN}, our proposed method consists of two parts. The upper part is the source network by macro-expression images and the bottom part is the target network by micro-expression images.

In source network, the input macro images are normalized in the pre-processing step. Macro images are fed into stochastic graph convolutional network. After each layer of stochastic graph convolution, an activation layer and pooling layer are appended. The classification is accomplished by computing the loss function at the fully connected layer. The SGCNs trained by source network are transferred to the target network.

In target network, there are two branches including spatial and temporal branch. The inputs of spatial and temporal branch are micro pixel images and optical flow images, respectively. In spatial branch, we use optical flow method to realize frame normalization to select meaningful frames in the entire micro-expression sequence. To improve the recognition performance, we adopt two transferred SGCNs (SGCN1 and SGCN2) to extract features, named dual SGCN (DSGCN). In temporal branch, we also utilize transferred SGCN1 and SGCN2 to extract features. Features from spatial and temporal branch are integrated to represent the discriminative features. The integrated features are fed into a fully connected layers.

\subsection{Stochastic Graph Construction}
\label{sec:SGCN}

We first give the detail of how our stochastic graph construction works. Given an image $I$, we convert it into an undirected and connected graph $\mathcal{G}=(\mathcal{V},\mathcal{E},\mathbf{A})$. $\mathcal{V}$ is a finite set of $|\mathcal{V}|=N$ vertices, where each node is corresponding to a pixel. $\mathcal{E}$ is a finite set of edges, which are decided by their neighbors. $\boldsymbol{x} \in \mathbb{R}^N$ represents the signal defined on the graph where
$x_i$ is the value of pixel at the $i^{th}$ node.

Our work presents a novel $p+q$ graph construction method. Each vertex has $p$ nearest neighbors and $q$ neighbors stochastically selected. The adjacency matrix $\mathbf{A}$ associated to $\mathcal{G}$ can be constructed. For node $i \in \{1 \cdots N\}$, we design a distance threshold $T$ to select its potential neighbors. We calculate the Euclidean distance $d(z_i,z_j)$ between node $i$ and node $j$, where $z_i$ is node $i$'s 2D coordinate. If $d_{ij}$ is not more than threshold $T$, node $j$ is a neighbor of node $i$, $j \sim i$. Otherwise, there is no edge between node $i$ and $j$. The weighs between node $i$ and node $j$ are calculated in (\ref{weight}) and the number of node $i$'s neighbors is denoted by $|\mathcal{N}_i|$.
\begin{equation}\label{weight}
    a_{ij}=\begin{cases}
    \text{exp}(-\frac{||z_i-z_j||_2^2}{\sigma^2}), &\text{if} \; d(z_i,z_j)\leq T \\
    0, &\text{otherwise}.
     \end{cases}
\end{equation}
where $z_i$ is the 2D coordinate of pixel $i$ and the $\sigma$ is the average distance between each vertex and the vertex that is the farthest from it.

Algorithm 1 gives the details of our adjacent matrix construction. For node $i$, we extract its potential neighbors $\mathcal{N}_i$ by (\ref{weight}) with given threshold $T$. We choose $p$ nearest neighbors as fixed neighbors. We compute the weights of $p$ nearest neighbors by Gaussian kernel function. We stochastically choose $q$ neighbors among the left.  The weights of remaining neighbors are zero.
\begin{table}[h]
    \footnotesize
    \justifying
    \begin{center}
    \begin{tabular}{ l }
    \toprule[1pt]
     \specialrule{0em}{1pt}{1pt}
      \textbf{Algorithm 1} Adjacency matrix of stochastic $p+q$ graph  \\  \specialrule{0em}{1pt}{1pt} \hline
     \textbf{Input:}  An image with $N$ vertices and node $i$'s 2D coordinate denoted \\
     \qquad \quad by $z_i$, $i=1 \cdots N$. The number of nearest neighbors denoted by\\
     \qquad \quad $p$ and stochastic neighbors denoted by $q$. The distance threshold \\
     \qquad \quad denoted by $T$. \\
     \textbf{Output:} Adjacency matrix $\mathbf{A}$\\
     \textbf{for} node $i=1$; $i\leq N$; $i++$  \textbf{do}\\
        \quad\; Step 1: Calculating the Euclidean distance $d(z_i,z_j)$ between node \\
        \quad\;  $i$ and node $j$. With threshold $T$, we can choose node $i$'s potential \\ \quad\;  neighbors $\mathcal{N}_i=\{i_1,i_2,\cdots, i_{|\mathcal{N}_i|}\}$. The weights between them\\
        \quad\; are calculated by (\ref{weight});\\
        \quad\; Step 2: Choosing $p$ nearest neighbors as fixed neighbors and\\ \quad\; stochastically choosing $q$ neighbors from the left, denoted these\\
        \quad\; $p+q$ neighbors by $\mathcal{S}$, where $\mathcal{S} \subset \mathcal{N}_i$. \\
        \quad\; Step 3: The weights of remaining neighbors are 0, where $j \notin \mathcal{S}$ \text{and}\\
         \quad\;\, $j \in \mathcal{N}_i/ \mathcal{S}$.  \\
     \textbf{end for}
    \\ \bottomrule[1pt]
    \end{tabular}
    \end{center}
    \label{Algorithm1}
\end{table}

\begin{figure}[htp]
   \centering
   \includegraphics[width=.45\linewidth]{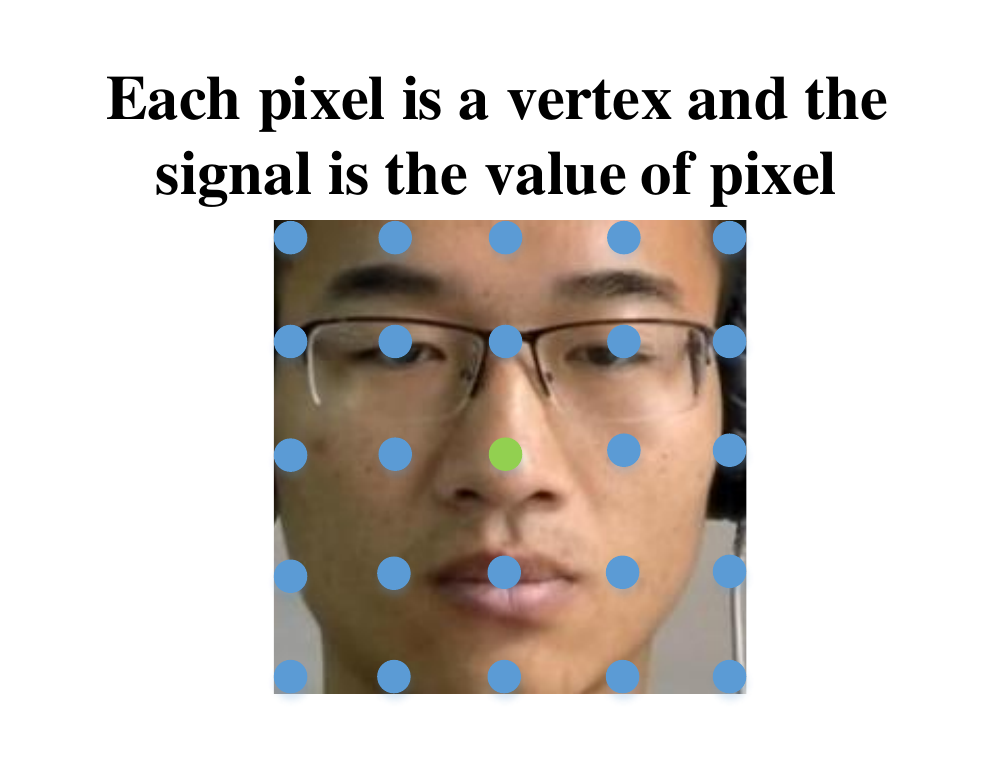}
   \includegraphics[width=.25\textwidth]{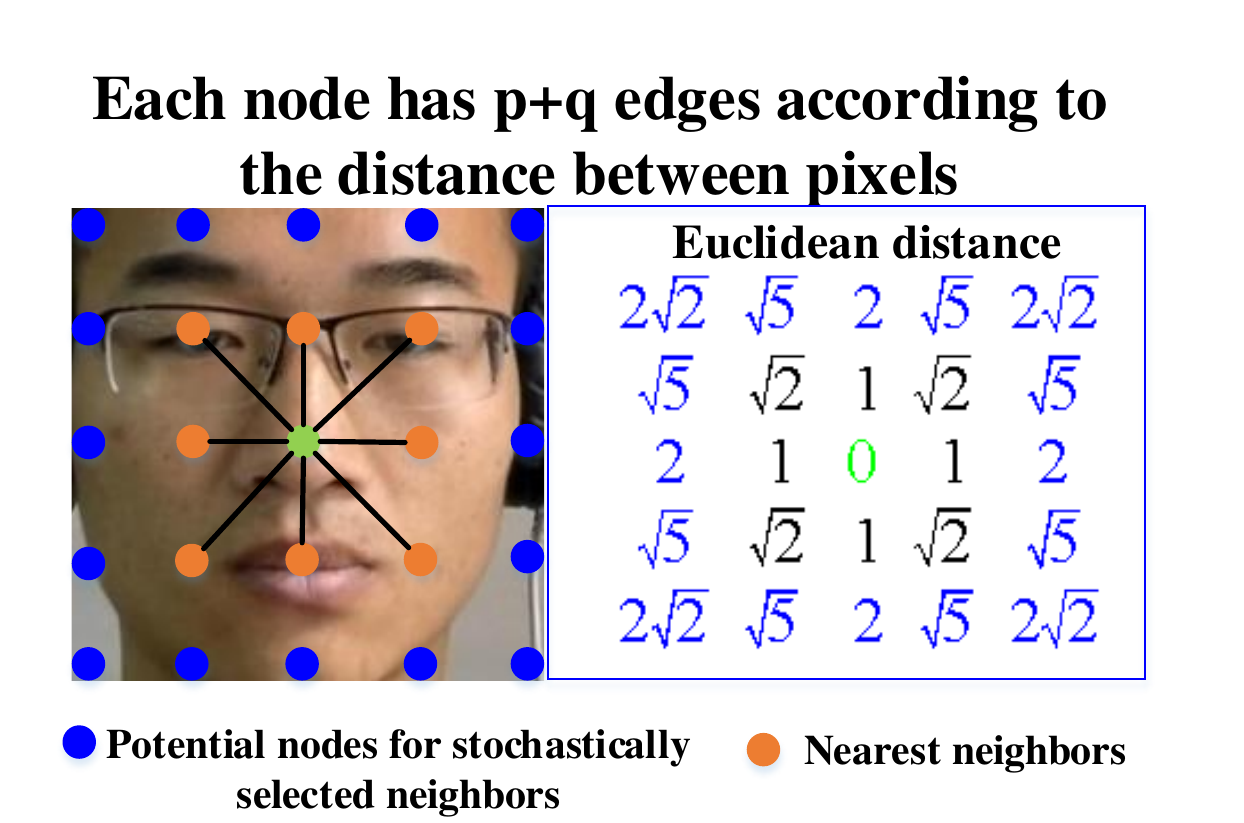}
   \caption{A simple example of $8+2$ graph construction. (\textbf{Left}): Node construction. (\textbf{Right}): Edge construction: the reference node (green) has 8 nearest neighbors with orange color and 2 neighbors stochastically selected from the potential nodes with blue color.}
  \label{graphcons}
\end{figure}

Figure \ref{graphcons} gives an example of $p=8$ and $q=2$ graph construction for the reference node with green color. We design the distance threshold $T=2\sqrt{5}$. There are 24 potential neighbors for the middle reference node. We choose 8 nearest neighbors as fixed neighbors. We stochastically select 2 neighbors from the left blue vertices. The remaining ones are unlinked.

\subsection{Dual Network and Transfer Learning }
We propose a dual network to integrate two different SGCNs (SGCN1 and SGCN2). SGCN1 uses the $p_1+q_1$ graph construction and SGCN2 uses $p_2+q_2$ graph construction. We use transfer learning to achieve feature extraction in our dual network. We use macro images to train the SGCN in source network. We use the trained model to extract features from micro images in the target network.

As illustrated in Section \ref{sec:SGCN}, we construct different adjacency matrices $\mathbf{A}_1$ and $\mathbf{A}_2$ corresponding to SGCN1 and SGCN2. The corresponding Laplacian matrices $\mathbf{L}_1$ and $\mathbf{L}_2$ are computed as
\begin{equation}
\begin{aligned}\label{Lap}
    \mathbf{L}_1 &= \mathbf{I}_N-\mathbf{D}_1^{-\frac{1}{2}}\mathbf{A}_1\mathbf{D}_1^{-\frac{1}{2}} \\
    \mathbf{L}_2 &= \mathbf{I}_N-\mathbf{D}_2^{-\frac{1}{2}}\mathbf{A}_2\mathbf{D}_2^{-\frac{1}{2}}
\end{aligned}
\end{equation}
where $\mathbf{D}_1$ and $\mathbf{D}_2$ are the diagonal degree matrices with $\mathbf{D}_1(i,i)=\sum\nolimits_{j}\mathbf{A}_1(i,j)$ and $\mathbf{D}_2(i,i)=\sum\nolimits_{j}\mathbf{A}_2(i,j)$. $\mathbf{I}_N \in \mathbb{R}^{N\times N}$ is the identity matrix.

There are $L$ stochastic graph convolutional layers in our dual network. We utilize the truncated Chebyshev polynomials to achieve the graph convolutional filtering \cite{NIPS2016}. For layer $l$, the output feature is denoted as $\mathbf{X}^{l} \in \mathbb{R}^{M \times d_l}$, where $M$ is the number of input samples and $d^l$ is the dimension of output features. The convolutional coefficients $\boldsymbol{w}^l\in \mathbb{R}^{K^l}$ are trained by macro images in source network,  where $K^l$ denotes the truncated order of Chebyshev polynomials in $l$-th layer. The dual stochastic graph convolution for the $l$-th layer ($1\leq l \leq L$) is implemented
\begin{equation}
\begin{aligned}\label{graphupdate}
    \mathbf{X}_1^{l}=\sum_{k=0}^{K^{l}-1}\boldsymbol{w}^1_k\mathbf{T}_k(\tilde{\mathbf{L}}_1)\mathbf{X}_1^{l-1}\\
    \mathbf{X}_2^{l}=\sum_{k=0}^{K^{l}-1}\boldsymbol{w}^2_k\mathbf{T}_k(\tilde{\mathbf{L}}_2)\mathbf{X}_2^{l-1}
\end{aligned}
\end{equation}
where $\boldsymbol{w}^1_k$ and $\textcolor{blue}{\boldsymbol{w}^2_k}$ are trained weights from the source network by macro images. $\tilde{\mathbf{L}}_1=\frac{2}{\lambda_{1_{max}}}\mathbf{L}_1-\mathbf{I}_N$ and $\lambda_{1_{max}}$ denotes the largest eigenvalues of $\mathbf{L}_1$. $\tilde{\mathbf{L}}_2=\frac{2}{\lambda_{2_{max}}}\mathbf{L}_2-\mathbf{I}_N$ and $\lambda_{2_{max}}$ denotes the largest eigenvalues of $\mathbf{L}_2$. The interval of the eigenvalues in $\tilde{\mathbf{L}}$ is $[-1,1]$.

We feed the outputs of $L$-th layer $\mathbf{X}^{L}_1 \in\mathbb{R}^{M \times d_L}$ and $\mathbf{X}^{L}_2 \in\mathbb{R}^{M \times d_L}$ into the trained fully connected layer. The outputs of fully connected layer are denoted as $\mathbf{X}^{fc}_1\in\mathbb{R}^{M \times d_{fc1}}$ and $\mathbf{X}^{fc}_2\in\mathbb{R}^{M \times d_{fc2}}$, where $d_{fc1}$ and $d_{fc2}$ are the dimensions of fully connected layer in SGCN1 and SGCN2, respectively.

After concatenation, the fused features of dual network are as follows
\begin{equation}\label{Dfeature}
\boldsymbol{X}^{D}=[\mathbf{X}^{fc}_1 \; \; \mathbf{X}^{fc}_2] \in \mathbb{R}^ {M \times (d_{fc1}+d_{fc2})}.
\end{equation}

\subsection{Features Extraction of Spatial and Temporal Branches}
In our proposed model, the spatial branch is designed to capture the spatial relationships between pixels across different frames. As we have mentioned in Section \ref{Intro}, optical flow images can provide temporal information between different frames. The spatial and temporal features of micro-expressions at different expression-states are encoded using DSGCNs.

\begin{figure}[htp]
   \centering
   \includegraphics[width=0.7\linewidth]{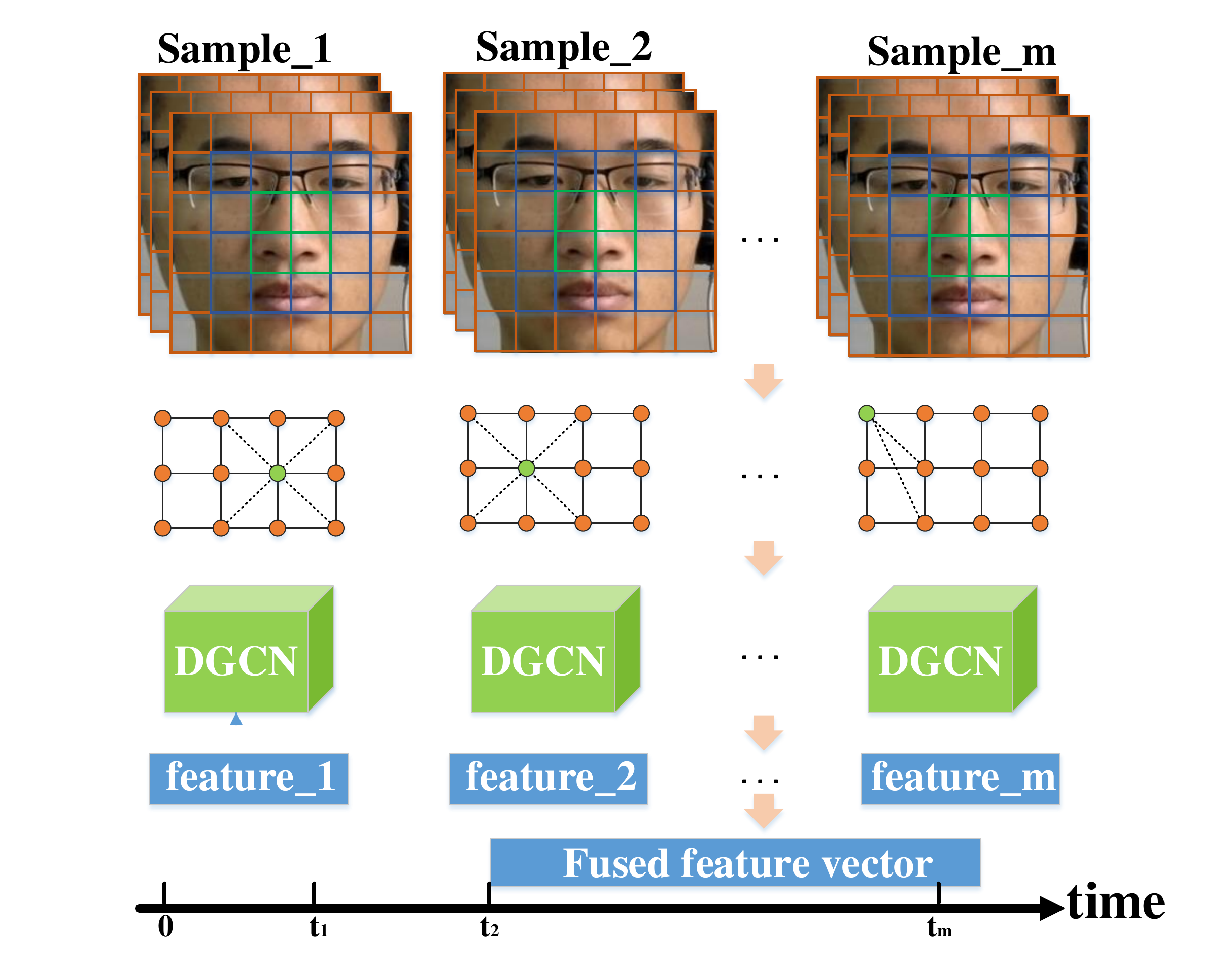}
   \caption{Illustration of feature extraction in the spatial branch}
  \label{spatialfeature}
\end{figure}

The feature extraction in the spatial branch is shown in Figure \ref{spatialfeature}. Specifically, given a video, the spatial branch of $m$-th frame is represented as $\mathcal{G}_m=(\mathcal{V}_m,\mathcal{E}_m,\mathbf{A}_m)$, where $m=1 \cdots M_s$ and $M_s$ is the number of image sequences selected from the video. As shown in Section \ref{sec:SGCN}, there are $N_s$ nodes in each image and the signal defined on the image is $\boldsymbol{x} \in \mathbb{R}^N_s$. The initial input features are represented by $\mathbf{X}_s^{in} = (\mathbf{x}_1, \mathbf{x}_2, \cdots, \mathbf{x}_{M_s}) \in \mathbb{R}^{M_s \times N_s}$, where $\mathbf{x}_m$ is the signal of $m$-th image. We feed the initial features $\mathbf{X}_s^{in}$ into the transferred DSGCN. By Eq. (\ref{Lap}), (\ref{graphupdate}) and (\ref{Dfeature}), we obtain the output features of spatial branch $\boldsymbol{X}_s^{D} \in  \mathbb{R}^{M_s \times (d_{fc1}+d_{fc2})}$.

\begin{figure}[htbp]
   \centering
   \includegraphics[width=0.7\linewidth]{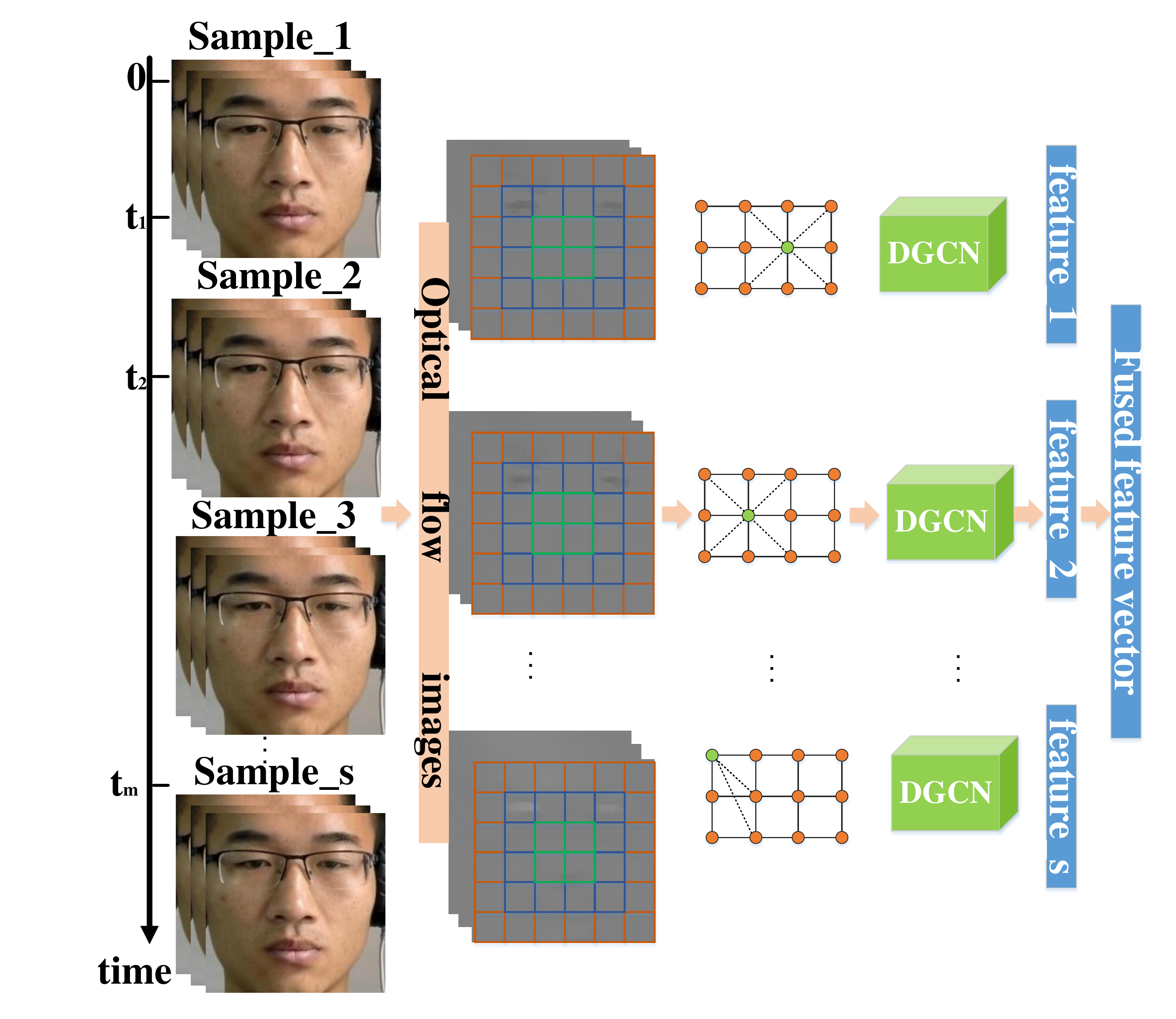}
   \caption{Illustration of feature extraction in the temporal branch}
  \label{temporalfeature}
\end{figure}

The feature extraction in temporal branch is shown in Figure \ref{temporalfeature}. We compute the optical flow features between sequences. We select $M_t$ samples from the optical flow sequences. The initial input features of temporal branch are represented by $\mathbf{X}_t^{in} \in \mathbb{R}^{M_t \times N_t}$, where $M_t$ is the number of pixels of optical flow image. We also feed the initial features $\mathbf{X}_t^{in}$ into the transferred DSGCN. By Eq. (\ref{Lap}), (\ref{graphupdate}) and (\ref{Dfeature}), we obtain the output features of temporal branch $\boldsymbol{X}_t^{D} \in  \mathbb{R}^{M_t \times (d_{fc1}+d_{fc2})}$.

We concatenate the features $\boldsymbol{X}_s^{D}$ and $\boldsymbol{X}_t^{D}$ obtained by spatial and temporal branches,
\begin{equation}\label{feature1}
    \boldsymbol{X}^{in}=[\boldsymbol{X}_s^{D} \\ \; \boldsymbol{X}_t^{D} ] \in \mathbb{R}^{(M_s+M_t) \times (d_{fc1}+d_{fc2})}
\end{equation}

We train a fully connected layer by the fused features $\boldsymbol{X}^{in}$ in target network. The output feature of the target network is denoted by  $\boldsymbol{X}^{out} \in \mathbb{R}^{(M_s+M_t) \times d_{fc3}} $, where $d_{fc3}$ is the dimension of fully connected layer appended by the spatial and temporal branch.

\subsection{Focal Loss Function}
Lin et al. \cite{He2017} proposed the definition of focal loss to handle the class imbalance in scene detection.  The focal loss is defined as:
\begin{equation}\label{Floss}
    FL(p_t)=-\alpha(1-p_t)^\gamma \log(p_t)
\end{equation}
\begin{equation}\label{Pt}
p_t=\left\{
             \begin{array}{lr}
             p & \text{if}  \ y=1 \\
             1-p & \text{otherwise}.
             \end{array}
\right.
\end{equation}
where $\alpha$ is a weighting factor to balance the importance for positive and negative example, $\gamma$ is the focusing parameters to balance the difference between easy and hard examples. $y\in \{0,1\}$ is the ground-truth label and $p\in[0,1]$ is the model's estimated probability for the class with label $y=1$.


\section{Experiments}

We set up a TensorFlow 1.14.0 environment on a Windows 10 computer with an i7-9700k processor and NVIDA GeForce RTX2070 graphics card.

\begin{table}[htb]
\centering
\footnotesize
\begin{tabular}{l|c}
\hline
  Layer & Parameters \\
\hline\hline
SGCN layer 1 & $K^1=9, d^1=32, \text{ReLU}, s_1=2$ \\
SGCN layer 2 & $K^2=9, d^2=32, \text{ReLU}, s_2=2$ \\
SGCN layer 3 & $K^3=6, d^3=64, \text{ReLU}, s_3=1$ \\
SGCN layer 4 & $K^4=6, d^4=64, \text{ReLU}, s_4=1$ \\
SGCN layer 5 & $K^5=4, d^5=128, \text{ReLU}, s_5=1$ \\
SGCN layer 6 & $K^6=4, d^6=128, \text{ReLU}, s_6=1$ \\
Fully connected &  $d_f=512, \text{ReLU}, \text{Dropout}$ \\
\hline
\end{tabular}
\caption{The network configuration of SGCN}
\label{NetPara}
\end{table}

\subsection{Datasets}
To evaluate our proposed approach, we conduct experiments using two databases: SAMM (the most commonly used) and MMEW (a newly published).

\textbf{MMEW} As a newly published dataset, there are $300$ samples in micro-and-macro expression warehouse (MMEW) \cite{Ben2021}, including happiness (36), anger (8), surprise (89), disgust (72), fear (16), sadness (13) and others (102), which is the biggest published micro-expression dataset. There are 36 participants including 9 females and 27 males, whose average age is 22.35. 900 macro-expression samples with the same category by the same group of participants are also provided.

\textbf{SAMM} The Spontaneous Actions and Micro-Movements
(SAMM) contains 159 samples
(image sequences containing spontaneous micro-expressions) recorded by a high-speed camera in a well controlled laboratory environment \cite{Davison2016, Davison2018, Yap2020}. There are 32 participants with a
mean age of 33.24 years, and an even male-female gender
split. Originally intended for investigating micro-facial
movements, the SAMM was induced based on the 7 basic
emotions including happiness (24), surprise (13), anger (20), disgust (8), sadness (3), fear (7), others (84).

\subsection{Experimental Settings}
As shown in \cite{Ben2021}, the performance results of published algorithms on micro-expression recognition are reported across different experimental settings. Ben \emph{et al}. conduct a fair comparison on datasets SAMM and MMEW, which are the most suitable datasets for recognition evaluation. We design the same experimental settings as \cite{Ben2021}. In the MMEW dataset, 226 samples from 5 classes (i.e., happiness, surprise, disgust, fear, sadness) were used; in the SAMM dataset,
72 samples from 5 classes (i.e., happiness, surprise, anger, disgust, fear) were used.

The network configuration of SGCN is designed in Table \ref{NetPara}. There are six stochastic convolutional layers. $K^l$ is the order of the ChebNet filter in the $l$-th layer , $d^l$ denotes the feature number of undirected graphs output from the $l$-th layer to the next layer, $s_l$ represents the number of times of graph coarsening in the $l$-th layer, $d_{fc}$ represents the size of fully connected layer.

The networks are trained using adaptive epochs or early stopping with a maximum set to 100 epochs. Basically, the training for each fold will stop when the loss score stops improving. We use Adaptive Moment Estimation (ADAM) as the optimizer, with a learning rate of $10^{-5}$ and decay of $10^{-6}$.

\begin{table}[htbp]
\centering
\footnotesize
\begin{tabular}{lcc}
  \hline
  \multirow{2}*{Methods} & \multicolumn{2}{c}{Recognition rate (\%)}\\ \cline{2-3}
                         & MMEW & SAMM \\ \hline\hline
  FDM \cite{Xu2017} & 34.6 & 34.1 \\
  ResNet10 \cite{HeK2016} & 36.6 & 39.3 \\
  Handcrafted features + deep learning \cite{Hu2018}& 36.6 & 47.1 \\
  LBP+TOP \cite{Zhao2007}& 38.9 & 37.0 \\
  Selective deep features \cite{Patel2017} & 39.0 & 42.9 \\
  ELRCN \cite{Khor2018} & 41.5 & 46.2 \\
  DCP-TOP \cite{Ben2018} & 42.5 & 36.8 \\
  ESCSTF \cite{Kim2016}  & 42.7 & 46.9 \\
  LHWP-TOP \cite{Ben2018} & 43.2 & 41.7 \\
  LBP-MOP \cite{Wang2015} & 43.9 & 35.3 \\
  LBP-SIP \cite{Wang2016} & 43.9 & 37.4 \\
  DiSTLBP-RIP \cite{Huang2017} & 44.0 & 46.2 \\
  RHWP-TOP \cite{Ben2018} & 45.9 &38.1 \\
  STLBP-IP \cite{Huang2015} & 46.6 & 42.9 \\
  ApexME \cite{Li2018}& 48.8 &50.0 \\
  Transfer Learning \cite{Peng2018} & 52.4 & 55.9 \\
  Multi-task mid-level feature learning \cite{He2017MER} & 54.2 & 55.0 \\
  KGSL \cite{Zong2018} & 56.9 & 48.6 \\
  Sparse MDMO \cite{Liu2018} & 60.0 & 52.9 \\
  MDMO \cite{Liu2016} & 65.7 & 50.0 \\
  DTSCNN \cite{Peng2017} & 65.9 & 69.2 \\
  TLCNN \cite{Wang2018} & 69.4 & 73.5 \\ \hline
  \textbf{Ours} & \textbf{72.7} & \textbf{75.0} \\\hline
\end{tabular}
\caption{Recognition rates (\%) of micro-expression using the state-of-the-art methods on MMEW and SAMM. The best record of each dataset is marked in bold.}
\label{statePer}
\end{table}

\begin{table*}[htbp]
\footnotesize
\centering
\begin{tabular}{l|ll|ll|ll}
    \hline
    \multirow{2}*{Experiments} &
    \multicolumn{2}{c|}{Pre-training}&
    \multicolumn{2}{c|}{Fine-tuning}&
    \multicolumn{2}{c}{Testing}\\
    \cline{2-7}
    & Data source & Rec.rate & Data source & Rec.rate & Data source & Rec.rate \\ \hline
    MMEW(Macro)$\rightarrow$ MMEW(Micro)  & MMEW (Macro) & 92.0\% & MMEW(Micro) & 96.6\%  & MMEW(Micro) & 69.4\% (\cite{Ben2021})  \\ \hline
    MMEW(Macro)$\rightarrow$ MMEW(Micro)  & MMEW (Macro) & 98.7\% & MMEW(Micro) & 99.3\%  & MMEW(Micro) & 72.7\% (\textbf{ours}) \\ \hline
    CK+  $\rightarrow$ SAMM & CK+ & 99.0\% &  SAMM & 99.7\% & SAMM & 73.5\% (\cite{Ben2021}) \\ \hline
    CK+  $\rightarrow$ SAMM & CK+ & 99.7\% &  SAMM & 97.5\% & SAMM & 75.0\% (\textbf{ours})\\ \hline
\end{tabular}
\caption{Data Sources and recognition rates of the pre-training , fine-tuning and testing results}
\label{MicroMac}
\end{table*}
\subsection{Comparison to State-of-the-art Methods}
To validate the effectiveness of our proposed method on the micro-expression recognition problem, we compare our
proposed method with several recent state-of-the-art methods on MMEW and SAMM. All the published results in \cite{Ben2021} are also kept for the convenience of comparison. Table \ref{statePer} summarizes the comparison
results.

It is obvious that all deep learning methods perform better than those utilizing handcrafted features. It can be seen that our proposed method achieves the best recognition performance ( 72.7\% on MMEW and
75.0\% on SAMM), which outperforms state-of-the-art method TLCNN
 ( the best recognition performance is 69.4\% on MMEW and
73.5\% on SAMM) \cite{Wang2018}.

The success of TLCNN \cite{Wang2018} demonstrates that the knowledge of macro-expressions is useful for micro-expression recognition under the CNN architecture. This paper explores the transferring macro-expression knowledge to assist micro-expression recognition based on the GCN architecture. The training and testing sets of MMEW and SAMM were set as \cite{Ben2021}. Table \ref{MicroMac}
lists the data source and accuracy of the pre-training, fine-tuning and testing results for each experiment. Compared to existing methods, our proposed method considers the stochastic geometric features of different pixels, which can provide more discriminative and robust information. These experimental results validate the superiority of our method.

\begin{table}
\centering
\footnotesize
\begin{tabular}{c|c|c|c|c|c}
  \hline
  \backslashbox{$\alpha$}{$\gamma$}& 0.1 & 0.2 & 0.5 & 1 & 1.5 \\ \hline
  $0.1$ & 54.55\% & 56.82\% & 54.55\% & 59.09\% & 56.82\% \\ \hline
  0.5 & 68.18\% & 70.45\% & 65.91\% & 68.18\% & 63.63\% \\ \hline
  $1$ & 68.18\% & 70.45\% & 70.45\% & 70.45\% & 70.45\% \\ \hline
  1.5 & 68.18\% & \textbf{72.73\%} & 65.19\% & 68.18\% & 65.91\% \\\hline
  2 & 68.18\% & 68.18\% & 68.18\% & 65.91\% & 68.18\% \\
  \hline
\end{tabular}
\caption{Comparison of the accuracies on MMEW with different focal loss parameters}
\label{focalMMEW}
\end{table}

\begin{figure}[htp]
   \centering
   \includegraphics[width=0.7\linewidth]{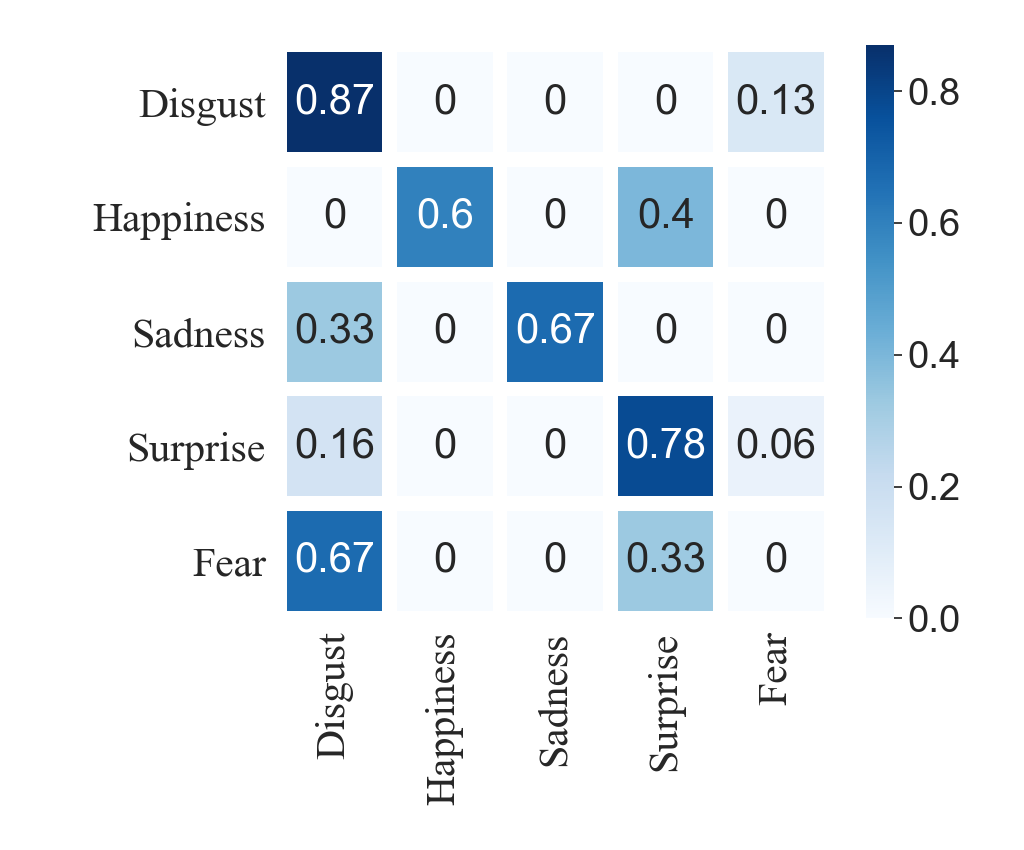}
   \caption{Confusion matrix of recognizing  5 expressions on MMEW dataset}
  \label{ConfMMEW}
\end{figure}
\textbf{Results on MMEW}
We design efficient parameters of focal loss on MMEW. The
results are presented in Table \ref{focalMMEW}. The best result is $72.73\%$ occurs with the parameters $\alpha = 1.5 $ and $\gamma = 0.2 $.
We also present confusion matrices in Figure \ref{ConfMMEW},
from which we observe that in MMEW, the “disgust”
and “surprise” samples can be highly recognized on
MMEW, instead the “fear” samples are difficult to train. Because the number of fear samples of MMEW is only 16, which is too small to train a good classifier.


\begin{table}
\footnotesize
\centering
\begin{tabular}{c|c|c|c|c|c}
  \hline
  \backslashbox{$\alpha$}{$\gamma$}& 0.1 & 0.2 & 0.5 & 1 & 1.5 \\ \hline
  $0.1$ & 68.75\% & 68.75\% & 62.5\% & 68.75\% & 56.25\% \\ \hline
  0.5 & 68.75\% & 68.75\% & 71.88\% & 68.75\% & 74.38\% \\ \hline
  $1$ & 68.75\% & 68.75\% & 74.38\% & \textbf{75.00\%} & 72.50\% \\ \hline
  1.5 & 71.88\% & 68.75\% & 68.75\% & 68.75\% & 68.75\% \\ \hline
  2 & 68.75\% & 68.75\% & 68.75\% & 68.75\% & 68.75\% \\
  \hline
\end{tabular}
\caption{Comparison of the accuracies on SAMM with different focal loss parameters}
\label{focalSAMM}
\end{table}

\begin{figure}[htp]
   \centering
   \includegraphics[width=0.7\linewidth]{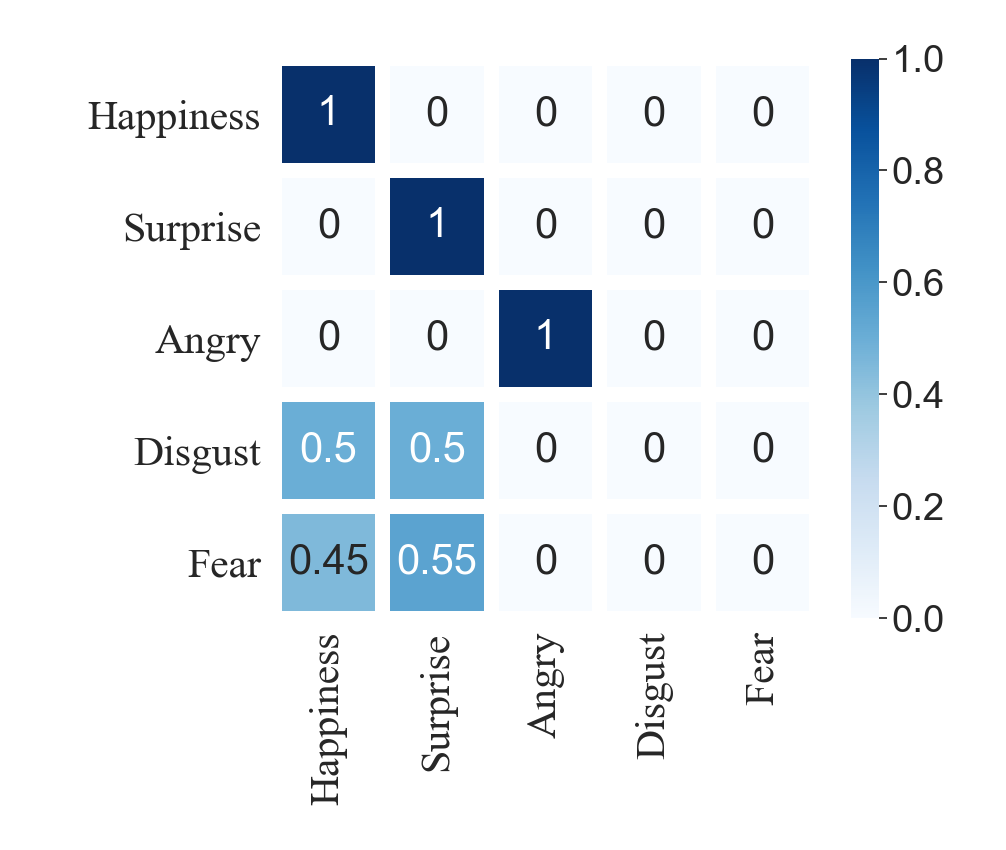}
   \caption{Confusion matrix of recognizing  5 expressions on SAMM dataset}
  \label{ConfSAMM}
\end{figure}

\textbf{Results on SAMM} We also design efficient parameters of focal loss on SAMM. The results of different weight factor $\alpha$ and focus factor $\gamma$ are presented in Table \ref{focalSAMM}. The best result $75.00\%$ occurs with the parameters
$\alpha = 1 $and $\gamma = 1 $.
The confusion matrix in Figure \ref{ConfSAMM} indicates that our method can totally recognize ”Happiness”, ”Surprise” and ”Angry”. It is difficult to train ”Disgust” and ”Fear” since their small samples.


\begin{table*}[htbp]
\footnotesize
\centering
\begin{tabular}{c|c|cc|cc|c|cc|cc}
    \hline
    \multirow{3}*{\makecell*[c]{Graph  \\ Architecture}} &
    \multicolumn{5}{c|}{MMEW}&
    \multicolumn{5}{c}{SAMM}\\
    \cline{2-11}
   & \multirow{2}*{Pre-training} & \multicolumn{2}{c|}{Spatial branch}&
    \multicolumn{2}{c|}{Temporal branch} & \multirow{2}*{Pre-training} & \multicolumn{2}{c|}{Spatial branch}&
    \multicolumn{2}{c}{Temporal branch} \\
    \cline{3-6}\cline{8-11}
   & & Fine-tuning & Testing & Fine-tuning & Testing & & Fine-tuning & Testing & Fine-tuning & Testing \\ \hline
    4+0   & 88.0\% & 93.1\% & 59.1\% & 40.9\% & 42.7\% & 99.8 \%  & 95.0\% & 62.5 \% & 35.5\% & 33.1\%\\ \hline
    4+6  & 88.0\% & 84.6\% & 55.2\% & 40.1\% & 41.4\% & 99.8 \%  & 97.5\% & 56.3 \% & 36.3\% & 32.5\% \\ \hline
    8+2   & 90.7\% & 93.7\% & 60.5\% & 40.3\% & 42.3\% & 98.9 \%  & 95.0\% & 56.3 \% & 35.0\% & 31.3\% \\ \hline
    12+0   & 93.3\% & 93.8\% & 61.8\% & 39.4\% & 40.5\% &  99.8 \%  & 97.2\% & 55.6 \% & 35.5\% & 32.5\% \\ \hline
    12+0 and 4+6 & 92.0\% & 96.5\% & 63.2\% & 40.0\% & 41.8\% & 98.4\%  & 97.5\% & 62.5 \% & 35.0\% & 31.3\% \\ \hline
    8+2 and 4+6   & 99.9\% & 96.6 \% &  63.4\% & 41.5\% & 42.5\% & 98.4\%  & 97.5\% & 56.3 \% & 36.0\% & 33.8\% \\ \hline
    8+2 and 12+0  & 98.7\% & 99.0\% & 63.6\% & 42.1\% & 42.3\% &   98.2\%  & 97.0\% & 62.5 \% & 35.5\% & 31.9\% \\ \hline
    4+0 and 12+0  & 98.7\% & 99.3\% & 60.7\% & 40.5\% & 42.5\% &  98.2 \%  & 97.5\% & 62.5 \% & 36.0\% & 34.4\% \\ \hline
    8+2 and 4+0  & 98.7\% & 97.4\% & \textbf{65.9\%} & 42.5\% & \textbf{45.7\%} &   98.0\%  & 97.3\% & \textbf{68.8 \%} & 38.8\% & \textbf{36.9\%} \\ \hline
    4+0 and 4+6  & 98.7\% & 96.0\% & 58.0\% & 41.5\% & 43.2\% &   98.0\%  & 97.5\% & 68.1 \% & 35.5\% & 33.1\% \\ \hline
\end{tabular}
\caption{Recognition rates comparison between dual and single graph convolutional network}
\label{DualGCN}
\end{table*}

\subsection{Ablation Study}
For further analysis, we perform an extensive ablation study by removing certain portions of our proposed TSDGCN to see how that affects performance. This was carried out using the databases MMEW and SAMM.

To verify the impact of DGCN, we train both single and dual GCN for the spatial branch and temporal branch under the same setting as Table \ref{MicroMac}. We select 10 sets of typical graph architectures containing 4 single GCN and 6 DGCN. For the convenience, we utilize the cross entropy loss here. The extensive experimental results on both MMEW and SAMM are reported in Table \ref{DualGCN}. It is obvious that most DGCNs obtain higher accuracies of pre-training, fine-tuning and testing than single GCNs. Comparing the testing accuracies, the best graph architecture is 8+2 and 4+0 on both MMEW and SAMM, which is marked in bold. We also observe that all the spatial branches have better recognition performance than temporal ones.

\begin{figure}[htp]
   \centering
   \includegraphics[width=1\linewidth]{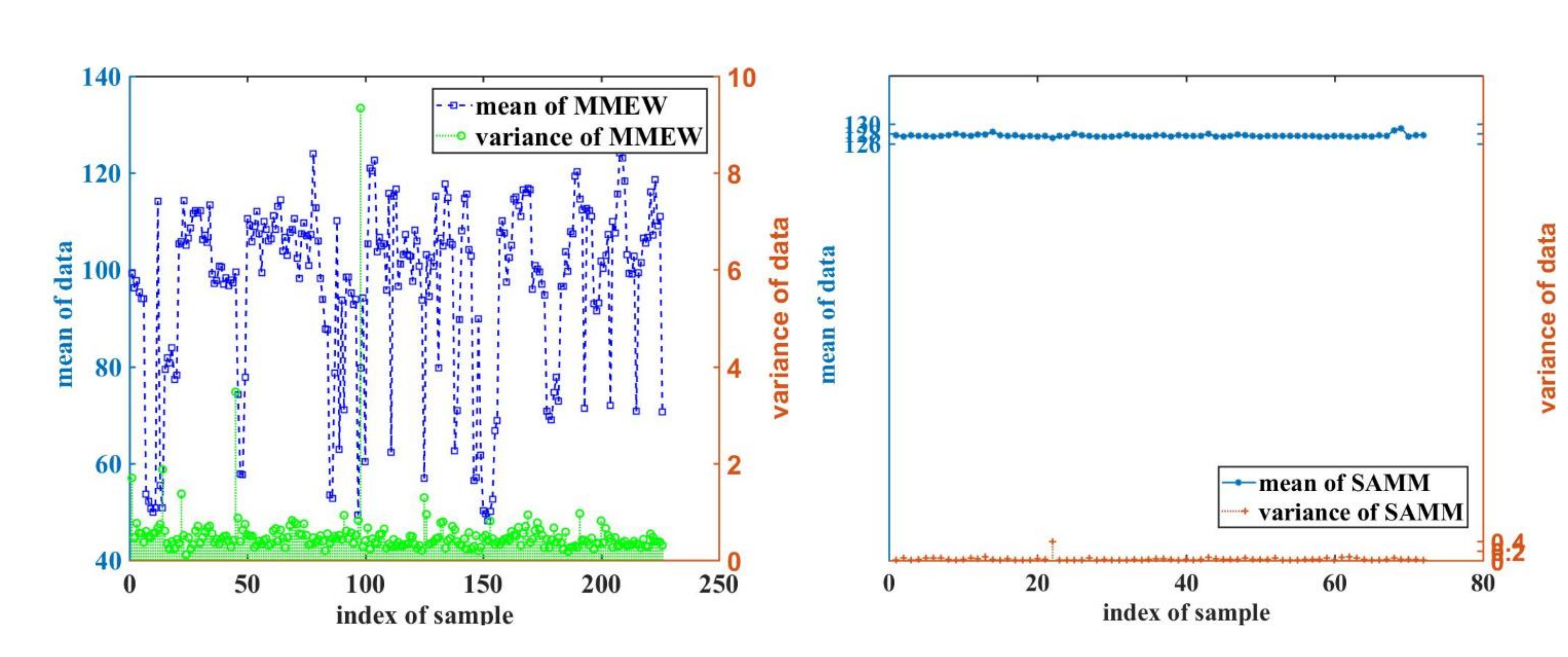}
   \caption{Mean and variance of optical flow sequences. (\textbf{Left}): MMEW. (\textbf{Right}): SAMM.}
   \label{optical}
\end{figure}

To verify the impact of spatial branch and the temporal branch, we train the ``Spatial branch" and ``Temporal branch" under the same experiment setting and graph architecture. With the aid of Table \ref{DualGCN}, we discuss the graph architecture 8+2 and 4+0. The experimental results on both MMEW and SAMM are reported in Table \ref{AlaStudy}. On MMEW, the recognition accuracy of ``Spatial +  CE" is 65.9\% and the recognition accuracy of ``Temporal +  CE" is only 45.7\%.  The performance of ``Spatial + Temporal branch + CE" varies with 8+2 and 4+0 graph architecture is the better than ``Spatial + CE" and ``Temporal branch + CE", which shows that ``Temporal branch" compensates the performance of ``Spatial branch". On SAMM, the recognition accuracy of ``Spatial +  CE" is 68.8\%  and the recognition accuracy of ``Temporal +  CE" is only 36.9\%. The performance of ``Spatial + Temporal branch + CE" is the same with ``Spatial+ CE", which means ``Temporal branch" doesn't improve the performance. Fig. \ref{optical} shows mean and variance of optical flow sequences on MMEW and SAMM. It is obvious that optical flow sequences of MMEW vary among samples and optical flow sequences of SAMM change a few. It means temporal branch  may not contain useful information, which is why temporal branch doesn't improve the performance in SAMM.

To verify the impact of focal loss, we compare the recognition performance between cross entropy and focal loss under the same architecture settings. ``Spatial+Temporal branch + FL" achieves the best performance on both MMEW and SAMM with the graph architecture 8+2 and 4+0, which shows the advantage of focal loss function.

\begin{table}[htbp]
\footnotesize 
\centering
\begin{tabular}{l|c|c|c}
    \hline
    \multirow{2}*{Method} & \multirow{2}*{Graph Architecture} & \multicolumn{2}{c}{Dataset}\\
      \cline{3-4} && MMEW&SAMM \\
    \hline
    Spatial + CE  & 8+2 and 4+0  & 65.9\% & 68.8 \% \\ \hline
    Temporal + CE  & 8+2 and 4+0   & 45.7\% & 36.9 \% \\ \hline
    Spatial + Temporal + CE  & 8+2 and 8+2 & 61.4\% & 68.8 \% \\ \hline
     Spatial + Temporal + CE  & 4+0 and 4+0  & 59.1\% & 68.8 \% \\ \hline
      Spatial + Temporal  + CE  & 8+2 and 4+0  & 70.5\% & 68.8 \% \\ \hline
     Spatial + Temporal + FL & 8+2 and 4+0 & \textbf{72.7}\% & \textbf{75.0} \% \\ \hline
\end{tabular}
\caption{Recognition rates of our proposed model with different graph architectures and loss functions.
“Spatial + CE” means we only feed pixel images into SGCNs and the loss is cross entropy. “Temporal” means we
we only feed optical flow images into SGCNs. "Spatial + Temporal" means we
we feed both pixel and optical flow images into SGCNs.  "FL" means the focal loss.}
\label{AlaStudy}
\end{table}

\subsection{Limitations}
Our proposed method may not have significant improved results when the dataset doesn't contain efficient temporal information.

\section{Conclusion}

We have proposed TSDGCN, a novel transferring stochastic dual graph convolutional network, which improves the recognition performance of facial micro-expression. We propose a novel method to construct a stochastic graph, which is computationally effecient. We present a dual GCN architecture to enhance the feature learning. We use large macro-expression datasets to train the SDCNs. We use the trained models to to extract the features of micro-expression samples. We extract and intergrade spatial and temporal features by SDGCNs. We introduce focal loss function to decrease the influence of class imbalance problem. Experimental results show that the proposed approach outperforms state-of-the-art methods. The idea of stochastic graph construction is a useful model can be extended to other graph application. 

{\small
\bibliographystyle{ieee_fullname}
\bibliography{egbib}

\begin{thebibliography}{10}\itemsep=-1pt

\bibitem{Atwood2016}
James Atwood and Don Towsley.
\newblock Diffusion-convolutional neural network.
\newblock {\em Adv. Neural Inform. Process. Syst.}, pages 2001--2009, 2016.

\bibitem{Ben2018}
Xianye Ben, Xitong Jia, Rui Yan, Xin Zhang, and Weixiao Meng.
\newblock Learning effective binary descriptors for micro-expression
  recognition transferred by macro-information.
\newblock {\em Pattern Recognition Letters}, 107:50--58, 2018.

\bibitem{Ben2021}
Xianye Ben, Yi Ren, Junping Zhang, Su jing Wang, Kidiyo Kpalma, Weixiao Meng,
  and Yong-Jin Liu.
\newblock Video-based facial micro-expression analysis: A survey of datasets,
  features and algorithms.
\newblock {\em IEEE Trans. Pattern Anal. Mach. Intell.}, Early access, 2021.

\bibitem{Bhushan2015}
Braj Bhushan.
\newblock Study of facial micro-expressions in psychology.
\newblock {\em Understanding facial expressions in communication}, pages
  265--286, 2015.

\bibitem{Bruna2014}
Joan Bruna, Wojciech Zaremba, Arthur Szlam, and Yann LeCun.
\newblock Spectral networks and locally connected networks on graphs.
\newblock {\em Int. Conf. Learn. Represent.}, 2014.

\bibitem{Chen2018}
Xinlei Chen, Li-Jia Li, Fei-Fei Li, and Abhinav Gupta.
\newblock Iterative visual reasoning beyond convolutions.
\newblock {\em IEEE Conf. Comput. Vis. Pattern Recog.}, pages 7239--7248, 2018.

\bibitem{Corneanu2016}
Ciprian~A. Corneanu, Marc Oliu, Jeffrey~F. Cohn, and Sergio Escalera.
\newblock Survey on {RGB}, {3D}, thermal, and multimodal approaches for facial
  expression recognition: History, trends, and affect-related applications.
\newblock {\em IEEE Trans. Pattern Anal. Mach. Intell.}, 38(8):1548--1568,
  2016.

\bibitem{Davision2015}
Adrian~K. Davision, Moi~Hoon Yap, and Cliff Lansley.
\newblock Micro-facial movement detection using individualised baselines and
  histogram based descriptors.
\newblock {\em IEEE International Conference on Systems, Man, and Cybernetics},
  pages 1864--1869, 2015.

\bibitem{Davison2016}
Adrian~K. Davison, Cliff Lansley, Nicholas Costen, Kevin Tan, and Moi~Hoon Yap.
\newblock Samm: A spontaneous micro-facial movement dataset.
\newblock {\em IEEE Transactions on Affective Computing}, 9(1):116--129, 2016.

\bibitem{Davison2018}
Adrian~K. Davison, Walied Merghani, and Moi~Hoon Yap.
\newblock Objective classes for micro-facial expression recognition.
\newblock {\em IEEE Transactions on Affective Computing}, 4(10):119, 2018.

\bibitem{NIPS2016}
Micha\"{e}l Defferrard, Xavier Bresson, and Pierre Vandergheynst.
\newblock Convolutional neural networks on graphs with fast localized spectral
  filtering.
\newblock {\em Adv. Neural Inform. Process. Syst.}, pages 3844--3852, 2016.

\bibitem{Ekman2003}
Paul Ekman.
\newblock Emotions revealed: Recognizing faces and feelings to improve
  communication and emotional life.
\newblock {\em Holt Paperback}, 128(8):140--140, 2003.

\bibitem{Ekman2009}
Paul Ekman.
\newblock Lie catching and microexpressions.
\newblock {\em The philosophy of deception}, pages 118--133, 2009.

\bibitem{Gan2019}
Y. Gan, Sze-Teng Liong, Wei-Chuen Yau, Yen-Chang Huang, and Lit-Ken Tan.
\newblock Off-apexnet on micro-expression recognition system.
\newblock {\em Signal Processing: Image Communication}, 74:129--139, 2019.

\bibitem{Gilmer2017}
Justin Gilmer, Samuel~S. Schoenholz, Patrick~F. Riley, Oriol Vinyals, and
  George~E. Dahl.
\newblock Neural message passing for quantum chemistry.
\newblock {\em Int. Conf. on Machine Learning}, pages 1263--1272, 2017.

\bibitem{He2017MER}
Jiachi He, Jian-Fang Hu, Xi Lu, and Wei-Shi Zheng.
\newblock Multi-task mid-level feature learning for micro-expression
  recognition.
\newblock {\em Pattern Recognition}, 66:44--52, 2017.

\bibitem{HeK2016}
Kaiming He, Xiangyu Zhang, Shaoqing Ren, and Jian Sun.
\newblock Deep residual learning for image recognition.
\newblock {\em IEEE Conf. Comput. Vis. Pattern Recog.}, pages 770--778, 2016.

\bibitem{Horn1981}
Berthold~K.P. Horn and Brian~G. Schunck.
\newblock Determining optical flow.
\newblock {\em Artificial Intelligence}, 17(1):185--203, 1981.

\bibitem{Hu2018}
Chunlong Hu, Dengbiao Jiang, Haitao Zou, Xin Zuo, and Yucheng Shu.
\newblock Multi-task micro-expression recognition combining deep and
  handcrafted features.
\newblock {\em IEEE International Conference on Pattern Recognition}, pages
  946--951, 2018.

\bibitem{Huang2017}
Xiaohua Huang, Sujing Wang, Xin Lin, Guoying Zhao, Xiaoyi Feng, and Matti
  Piteikainen.
\newblock Discriminative spatio temporal local binary pattern with revisited
  integral projection for spontaneous facial micro-expression recognition.
\newblock {\em IEEE Transactions on Affective Computing}, 10:32--47, 2017.

\bibitem{Huang2015}
Xiaohua Huang, Sujing Wang, Guoying Zhao, and Matti Piteikainen.
\newblock Facial microexpression recognition using spatiotemporal local binary
  pattern with integral projection.
\newblock {\em Int. Conf. Comput. Vis.}, pages 1--9, 2015.

\bibitem{Ji2020}
Sijie Ji, Kai Wang, Xiaojiang Peng, Jianfei Yang, Zhaoyang Zeng, and Yu Qiao.
\newblock Multiple transfer learning and multi-label balanced training
  strategies for facial \text{AU} detection in the wild.
\newblock {\em IEEE Conf. Comput. Vis. Pattern Recog.}, pages 1657--1661, 2020.

\bibitem{Khor2018}
Huaiqian Khor, John See, Raphael C.W.Phan, and Weiyao Lin.
\newblock Enriched long-term recurrent convolutional network for facial
  micro-expression recognition.
\newblock {\em IEEE International Conference on Automatic Face and Gesture
  Recognition}, pages 234--778, 2018.

\bibitem{Kim2016}
Dae~Hoe Kim, Wissam~J. Baddar, and Yong~Man Ro.
\newblock Micro-expression recognition with expression-state constrained
  spatio-temporal feature representations.
\newblock {\em ACM Int. Conf. Multimedia}, pages 382--386, 2016.

\bibitem{Kipf2017}
Thomas~N. Kipf and Max Welling.
\newblock Semi-supervised classification with graph convolutional networks.
\newblock {\em Int. Conf. Learn. Represent.}, 2017.

\bibitem{Landrieu2018}
Loic Landrieu and Martin Simonovsky.
\newblock Large-scale point cloud semantic segmentation with superpoint graphs.
\newblock {\em IEEE Conf. Comput. Vis. Pattern Recog.}, pages 4558--4567, 2018.

\bibitem{Li2017}
Shan Li and Weihong Deng.
\newblock Reliable crowdsourcing and deep locality-preserving learning for
  facial expression in the wild.
\newblock {\em IEEE Conf. Comput. Vis. Pattern Recog.}, pages 2584--2593, 2017.

\bibitem{Li2019}
Shan Li and Weihong Deng.
\newblock Reliable crowdsourcing and deep locality-preserving learning for
  unconstrained facial expression recognition.
\newblock {\em IEEE Trans. Image Process.}, 28(1):356--370, 2019.

\bibitem{Li2018}
Yante Li, Xiaohua Huang, and Guoying Zhao.
\newblock Can micro-expression be recognized based on single apex frame $?$.
\newblock {\em IEEE International Conference on Image Processing}, pages
  3094--3098, 2018.

\bibitem{He2017}
Tsung-Yi Lin, Priya Goyal, Ross Grishick, Kaiming He, and Piotr
  Doll$\acute{a}$r.
\newblock Focal loss for dense object detection.
\newblock {\em Int. Conf. Comput. Vis.}, pages 2980--2988, 2017.

\bibitem{Liu2018}
Yong-Jin Liu, Bing-Jun Li, and Yu-Kun Lai.
\newblock Sparse {MDMO}: Learning a discriminative feature for spontaneous
  micro-expression recognition.
\newblock {\em IEEE Transactions on Affective Computing}, 12(1):254--261, 2018.

\bibitem{Liu2016}
Yong-Jin Liu, Jin-Kai Zhang, Wen-Jing Yan, Su-Jing Wang, Guoying Zhao, and
  Xiaolan Fu.
\newblock A main directional mean optical flow feature for spontaneous
  micro-expression recognition.
\newblock {\em IEEE Transactions on Affective Computing}, 7(4):299--310, 2016.

\bibitem{Lo2020}
Ling Lo, Hong-Xia Xie, Hong-Han Shuai, and Wen-Huang Cheng.
\newblock {MER-GCN}: Micro-expression recognition based on relation modeling
  with graph convolutional networks.
\newblock {\em IEEE Conference on Multimedia Information Processing and
  Retrieval}, pages 79--84, 2020.

\bibitem{Alessio2009}
Alessio Micheli.
\newblock Neural network for graphs: a contextual constructive approach.
\newblock {\em IEEE Transactions on Neural Networks}, 20(3):498--511, 2009.

\bibitem{Ali2019}
Ali Mollahosseini, Behzad Hasani, and Mohammad~H. Mahoor.
\newblock Affectnet: A database for facial expression, valence, and arousal
  computing in the wild.
\newblock {\em IEEE Transactions on Affective Computing}, 10:18--31, 2019.

\bibitem{Niepert2016}
Mathias Niepert, Don~Mohamed Ahmed, and Konstantin Kutzkov.
\newblock Learning convolutional neural networks for graphs.
\newblock {\em Int. Conf. on Machine Learning}, pages 2014--2023, 2016.

\bibitem{Ojala2002}
Timo Ojala, Matti Pietikainen, and Topi Maenpaa.
\newblock Multiresolution gray-scale and rotation invariant texture
  classification with local binary patterns.
\newblock {\em IEEE Trans. Pattern Anal. Mach. Intell.}, 24(7):971--987, 2002.

\bibitem{Patel2017}
Davangini Patel, Xiaopeng Hong, and Guoying Zhao.
\newblock Selective deep features for micro-expression recognition.
\newblock {\em IEEE International Conference on Pattern Recognition}, pages
  2258--2263, 2017.

\bibitem{Peng2017}
Min Peng, Chongyang Wang, Tong Chen, Guangyuan Liu, and Xiaolan Fu.
\newblock Dual temporal scale convolutional neural network for micro-expression
  recognition.
\newblock {\em Frontiers in Psychology}, 8:1745--1756, 2017.

\bibitem{Peng2018}
Min Peng, Zhan Wu, Zhizhao Zhang, and Tong Chen.
\newblock From macro to micro expression recognition: deep learning on small
  datasets using transfer learning.
\newblock {\em IEEE International Conference on Automatic Face and Gesture
  Recognition}, pages 657--661, 2018.

\bibitem{Chaudhry2009}
Avinash~Ravichandran Riawan~Chaudhry, Gregory Hager, and Rene Vidal.
\newblock Histograms of oriented optical flow and binet-cauchy kernels on
  nonlinear dynamical systems for the recognition of human actions.
\newblock {\em IEEE Conf. Comput. Vis. Pattern Recog.}, pages 1932--1939, 2009.

\bibitem{Wang2018}
Sujing Wang, Bingjun Li, Yongjin Liu, Wenjing Yan, Xinyu Ou, Xiaohua Huang,
  Feng Xu, and Xiaolan Fu.
\newblock Micro-expression recognition with small sample size by transferring
  long-term convolutional neural network.
\newblock {\em Neurocomputing}, 312:251--262, 2018.

\bibitem{Wang2015TIP}
Sujing Wang, Wen-Jing Yan, Xiaobai Li, Guoying Zhao, Chun-Guang Zhou, Xiaolan
  Fu, Minghao Yang, and Jianhua Tao.
\newblock Micro-expression recognition using color spaces.
\newblock {\em IEEE Trans. Image Process.}, 24(12):6034--6047, 2015.

\bibitem{Wang2014ECCV}
Sujing Wang, Wen-Jing Yan, Guoying Zhao, Xiaolan Fu, and Chunguang Zhou.
\newblock Micro-expression recognition using robust principal component
  analysis and local spatiotemporal directional features.
\newblock {\em Eur. Conf. Comput. Vis.}, pages 325--338, 2014.

\bibitem{Wang2015}
Yangan Wang, John See, Raphael C.-W. Phan, and Yee-Hui Oh.
\newblock Efficient spatio-temporal local binary patterns for spontaneous
  facial micro-expression recognition.
\newblock {\em PLoS ONE}, 10(5):e0124674, 2015.

\bibitem{Wang2016}
Yangan Wang, John See, Raphael C.-W. Phan, and Yee-Hui Oh.
\newblock {LBP} with six intersection points: Reducing redundant information in
  {LBP-TOP} for micro-expression recognition.
\newblock {\em Asian Conf. on Comput. Vis.}, pages 382--386, 2016.

\bibitem{Pan2021}
Zonghan Wu, Shirui Pan, Fengwen Chen, Guodong Long, Chengqi Zhang, and Philip~S
  Yu.
\newblock A comprehensive survey on graph neural networks.
\newblock {\em IEEE Transactions on Neural Networks and Learning Systems},
  32(1):4--24, January 2021.

\bibitem{Xia2020}
Zhaoqiang Xia, Wei Peng, Huaiqian Khor, Xiaoyi Feng, and Guoying Zhao.
\newblock Revealing the invisible with model and data shrinking for
  composite-database micro-expression recognition.
\newblock {\em IEEE Trans. Image Process.}, 29:8590--8605, 2020.

\bibitem{Xie2020}
Hong-Xia Xie, Ling Lo, Hong-Han Shuai, and Wen-Huang Cheng.
\newblock {AU}-assisted graph attention convolutional network for
  micro-expression recognition.
\newblock {\em ACM Int. Conf. Multimedia}, pages 79--84, 2020.

\bibitem{Xu2017}
Feng Xu, Junping Zhang, and James~Z. Wang.
\newblock Micro-expression identification and categorization using a facial
  dynamics map.
\newblock {\em IEEE Transactions on Affective Computing}, 8(2):254--267, 2017.

\bibitem{Yap2020}
Chuin~Hong Yap, Connah Kendrick, and Moi~Hoon Yap.
\newblock Samm long videos: a spontaneous facial micro- and macro-expressions
  dataset.
\newblock {\em IEEE International Conference on Automatic Face and Gesture
  Recognition}, pages 771--776, 2020.

\bibitem{Zach2007}
C. Zach, T. Pock, and H. Bischof.
\newblock A duality based approach for realtime {TV-L1} optical flow.
\newblock {\em Pattern Recognition}, pages 214--223, 2007.

\bibitem{Zhao2007}
Guoying Zhao and Matti Paetikainen.
\newblock Dynamic texture recognition using local binary patterns with an
  application to facial expressions.
\newblock {\em IEEE Trans. Pattern Anal. Mach. Intell.}, 29(6):915--928, 2007.

\bibitem{Zong2018}
Yuan Zong, Xiaohua Huang, Wenming Zheng, Zhen Cui, and Guoying Zhao.
\newblock Learning from hierarchical spatiotemporal descriptors for
  microexpression recognition.
\newblock {\em IEEE Trans. Multimedia}, 20(11):3160--3172, 2018.

\bibitem{Zong2018TL}
Yuan Zong, Wenming Zheng, Xiaohua Huang, Jingang Shi, Zhen Cui, and Guoying
  Zhao.
\newblock Domain regeneration for cross-database micro-expression recognition.
\newblock {\em IEEE Trans. Image Process.}, 27(5):2484--2498, 2018.

\bibitem{Zong2019}
Yuan Zong, Wenming Zheng, ZhenCui, Guoying Zhao, and Bin Hu.
\newblock Toward bridging micro-expressions from different domain.
\newblock {\em IEEE Transactions on Cybernetics}, 50(12):5047--5060, 2019.

\end{thebibliography}
}

%
%

\end{document}